\documentclass[letterpaper, 10 pt, conference]{ieeeconf}  

\IEEEoverridecommandlockouts                              

\usepackage{kotex}
\usepackage{dsfont}
\usepackage{amsfonts}
\usepackage{amsmath}
\usepackage{amssymb}
\usepackage{textcomp}
\usepackage{multirow}
\usepackage{graphicx}
\usepackage{lipsum}
\usepackage{longtable}
\usepackage{booktabs}
\usepackage{dblfloatfix}
\usepackage{xcolor} 
\usepackage{hyperref}
\usepackage{xfrac}    
\usepackage{algpseudocode}
\usepackage{lipsum} 
\usepackage{wrapfig}
\usepackage{multicol}
\usepackage{bbm}
\usepackage[ruled,vlined]{algorithm2e}
\usepackage{pifont}
\newcommand{\cmark}{\ding{51}}%

\usepackage{tabularx} 

\title{\LARGE \bf
DGS-SLAM: Gaussian Splatting SLAM in Dynamic Environment
}

\author{Mangyu Kong$^{1}$, Jaewon Lee$^{1}$, Seongwon Lee$^{2}$ and Euntai Kim$^{1}$
\thanks{$^{1}$M.\,Kong, J.\,Lee and E.\,Kim are with the School of Electrical and Electronic Engineering,  Yonsei University, Seoul 03722, South Korea, {\tt\small \{mangyu0929,leejaewon,etkim\}@yonsei.ac.kr}}
\thanks{$^{2}$S.\,Lee is with the School of Electrical Engineering, Kookmin University, Seoul 02707, South Korea, {\tt\small sungonce@kookmin.ac.kr}}%
}

\begin{document}

\maketitle
\thispagestyle{empty}
\pagestyle{empty}

\begin{abstract}

We introduce Dynamic Gaussian Splatting SLAM (DGS-SLAM), the first dynamic SLAM framework built on the foundation of Gaussian Splatting. While recent advancements in dense SLAM have leveraged Gaussian Splatting to enhance scene representation, most approaches assume a static environment, making them vulnerable to photometric and geometric inconsistencies caused by dynamic objects. To address these challenges, we integrate Gaussian Splatting SLAM with a robust filtering process to handle dynamic objects throughout the entire pipeline, including Gaussian insertion and keyframe selection. Within this framework, to further improve the accuracy of dynamic object removal, we introduce a robust mask generation method that enforces photometric consistency across keyframes, reducing noise from inaccurate segmentation and artifacts such as shadows.
Additionally, we propose the loop-aware window selection mechanism, which utilizes unique keyframe IDs of 3D Gaussians to detect loops between the current and past frames, facilitating joint optimization of the current camera poses and the Gaussian map. DGS-SLAM achieves state-of-the-art performance in both camera tracking and novel view synthesis on various dynamic SLAM benchmarks, proving its effectiveness in handling real-world dynamic scenes. The codes are available at \href{https://github.com/kmk97/DGS-SLAM}{this url}.

\end{abstract}

\section{Introduction}

Dense Simultaneous Localization and Mapping (SLAM) has been a fundamental task in robotics and computer vision for decades. The primary goal of dense SLAM is to simultaneously estimate the camera’s position and reconstruct a dense map in an unknown environment. This task is crucial for many real-world applications such as robot navigation, autonomous driving, and VR/AR. Since the specific algorithm used in dense SLAM is inherently linked to the type of map representation, the choice of representation has a significant impact on SLAM performance. As a result, various approaches have been explored, including surfels~\cite{dense_surfel,elasticfusion}, meshes~\cite{surfelmeshing}, and signed distance fields (SDF)~\cite{kinectfusion,refusion}.

Recently, NeRF-based volumetric representation~\cite{nerf,instantngp} has emerged as a promising map representation, offering detailed scene reconstruction. These methods capture dense photometric information from the environment, enabling high-quality image synthesis from novel viewpoints. Following these advancements, Gaussian Splatting~\cite{3dgs} stands out as a novel volumetric technique. Unlike ray-matching methods like NeRF~\cite{nerf}, Gaussian Splatting employs differential rasterization of 3D primitives, significantly enhancing reconstruction quality and rendering speed.

\begin{figure}[htbp]
\centering
\includegraphics[width=0.9\columnwidth]{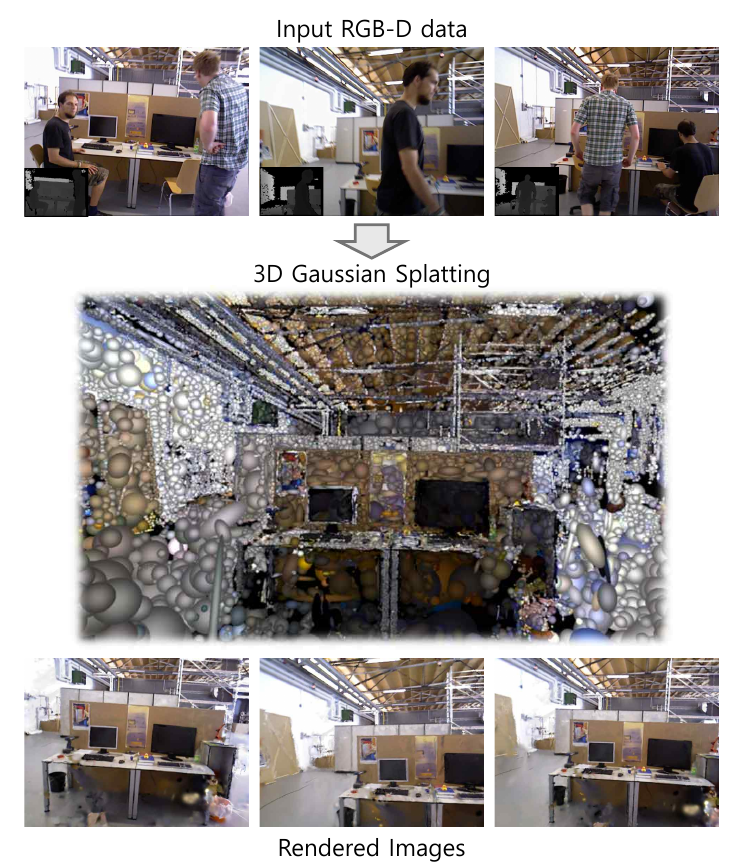}
\vspace{-12pt}
\caption{Result of our framework. Top: RGB-D frames as input. Center: Reconstructed Gaussian map model without dynamics. Bottom: Rendered images from tracked camera pose.}
\vspace{-15pt}
\label{fig:fig1} 
\end{figure}

With these capabilities, various SLAM methods utilizing Gaussian Splatting~\cite{keetha2024splatam,matsuki2024gaussian,ha2024rgbd,photo,compact} have been introduced, demonstrating remarkable improvements in SLAM performance. Gaussian Splatting enables highly efficient and accurate map representations with reduced computational overhead, making it a compelling choice for SLAM applications. Despite its advantages, however, most existing methods assume static environments, leading to inaccuracies and degraded performance in dynamic scenarios. Therefore, addressing this constraint is essential for expanding the applicability of Gaussian Splatting SLAM to real-world scenarios.

On the other hand, several traditional SLAM methods have been developed to address the challenges posed by complex dynamic environments. These approaches utilize techniques such as semantic segmentation priors~\cite{ds_slam,dynaslam}, optical flow~\cite{flowfusion,sun2018motion}, and residual optimization~\cite{refusion} to filter out moving objects. While these approaches have shown some success in mitigating the effects of dynamic elements, they also come with limitations. Methods that rely on semantic priors often struggle with segmentation errors and artifacts in real-world scenarios. Although residual optimization can be effective, it tends to fail when confronted with large object movements. Moreover, traditional dynamic SLAM systems are generally limited in their ability to generate detailed scene representations.

To address both the challenges of dynamic environments and the limitations of traditional dynamic SLAM methods, we introduce Dynamic Gaussian Splatting SLAM (DGS-SLAM) in this work.
Our DGS-SLAM is designed to integrate Gaussian Splatting SLAM with a robust dynamic filtering process to handle dynamic objects throughout the entire SLAM pipeline, from Gaussian initialization to joint optimization. We fully exploit the inherent properties of Gaussian Splatting to ensure both robustness and efficiency in dynamic environments.

In this framework, we additionally propose a robust mask generation method that improves the accuracy of dynamic object removal by ensuring photometric consistency across keyframes. This method helps to reduce noise from inaccurate segmentation and minimize the impact of artifacts such as shadows, reliably separating dynamic and static elements in the scene. We also propose a Gaussian-based loop-aware window selection strategy utilizing unique keyframe IDs associated with each Gaussian. This enables joint optimization across previous keyframes, further enhancing the framework's ability to maintain accurate localization over time while revisiting scenes in dynamic environments.

Our proposed DGS-SLAM achieves state-of-the-art performance in both camera tracking and novel view synthesis on various dynamic SLAM benchmarks, proving its effectiveness in handling real-world dynamic scenes.

\vspace{-1mm}
\section{Related Works}

\begin{figure*}[htbp]
\centering
\includegraphics[width=\textwidth]{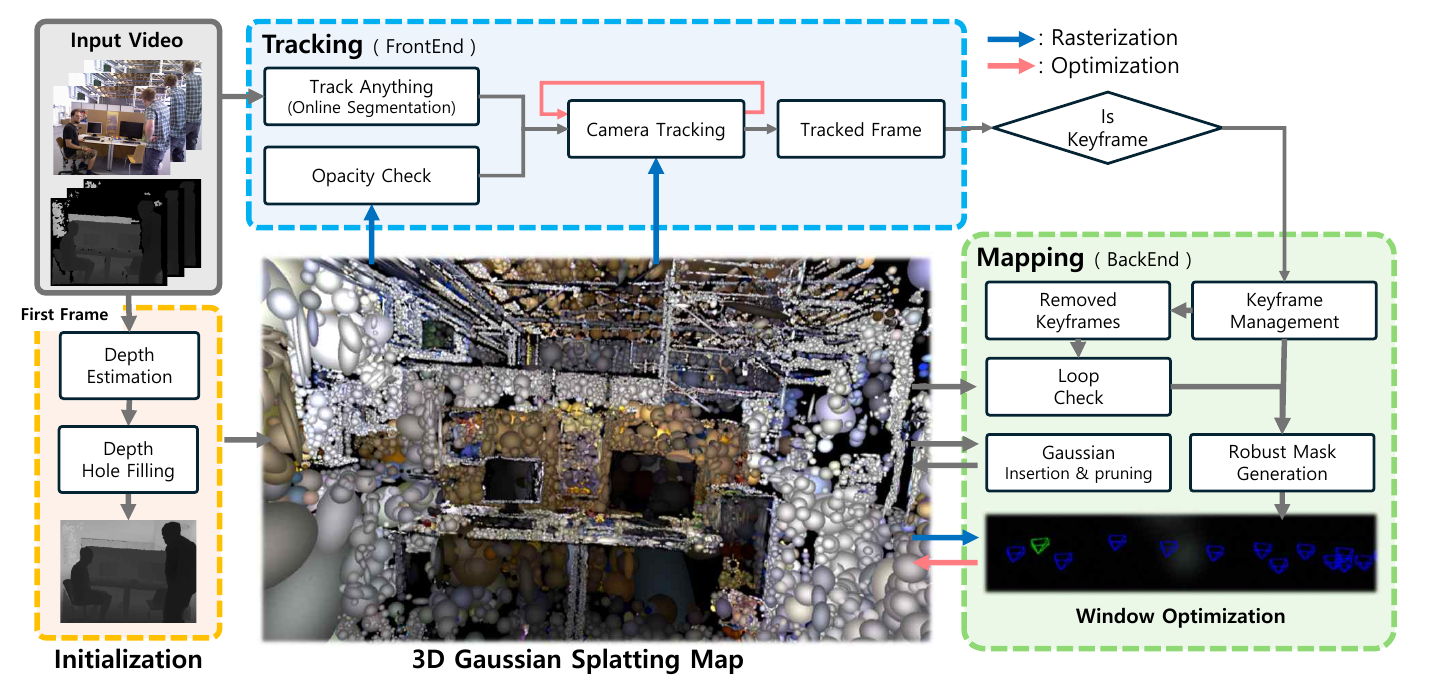}
\vspace{-7pt}
\caption{\textbf{Framework Overview} Our framework simultaneously estimates the camera pose while reconstructing a 3D gaussian splatting map with a sequence of RGB-D frames in a dynamic environment. DGS-SLAM consists of three main components: initialization, frontend tracking, and backend mapping. During initialization, the Gaussians are optimized based on the first frame. In the frontend, DGS-SLAM estimates the camera pose while filtering out dynamic elements. The backend then performs joint optimization to refine the pose and update the 3D Gaussian Splatting map.}
\vspace{-12pt}
\label{fig:fig2} 
\end{figure*}
\subsection{Dynamic SLAM}

Dynamic SLAM focuses on accurately estimating poses and reconstructing static scenes in environments where objects are moving or changes occur. Various dynamic SLAM methods have been developed to address these challenges. Some approaches use deep learning segmentation modules to remove dynamic elements from the scene, enabling accurate pose prediction and mapping~\cite{ds_slam,dynaslam,vdoslam}. Others utilize dense optical flow module~\cite{raft,gmflow} to detect motion and identify dynamic regions~\cite{flowfusion,sun2018motion,cheng2019improving}. Additionally, some methods leverage residuals obtained after initial registration to filter out outliers~\cite{refusion}. However, these methods often suffer from noise due to domain gaps in real-world environments and struggle with large movements of dynamic objects. Moreover, in existing dense dynamic SLAM, achieving photorealistic rendering is challenging. In this work, we employ Gaussian Splatting as a map representation in dynamic SLAM, enabling high-quality rendering from a novel view. Furthermore, we propose a robust mask generation methods to complement inaccurate segmentations and artifacts from dynamic objects.

\subsection{Neural Implicit SLAM}

Neural implicit representations like NeRF~\cite{nerf} and SDF~\cite{instantngp}, have gained significant attention for its remarkable ability to represent scenes densely and continuously. iMap~\cite{imap}, the first approach to apply neural implicit representation in SLAM, achieved real-time mapping and tracking. After iMAP, Nice-SLAM~\cite{nice_slam} introduced a hierarchical multi-feature representation to improve scalability and address over-smoothed scene reconstruction. Following these, several studies~\cite{voxfusion,eslam,co-slam,point-slam,vmap} propose more practical scene representations to address memory limitations and enhance reconstruction accuracy. While most neural implicit SLAM systems demonstrate impressive performance under the assumption of static scenes, they show limitations in dynamic environments. Our DGS-SLAM addresses challenges in dynamic environments and demonstrates outstanding performance by utilizing a novel map representation.

\subsection{3D Gaussian Splatting SLAM}

Gaussian Splatting~\cite{3dgs} has recently emerged as an effective method for representing scenes through a set of Gaussians. Unlike neural implicit representations that rely on ray marching, 3D Gaussian Splatting utilizes differential rasterization, achieving both fast rendering speeds and accurate scene reconstruction. Leveraging these advantages, this approach has expanded into various fields, including more accurate 3D scene reconstruction~\cite{2dgs,scaffold,mipgs}, dynamic scene modeling~\cite{4dgs,deformable,gaussianflow,spacetime}, and scene editing~\cite{featuregs,gaussianeditor}. One of the most developed areas is Dense SLAM utilizing Gaussian Splatting as map representation. SplaTAM~\cite{keetha2024splatam} introduced silhouette-guided optimization for progressive map reconstruction in dense SLAM. Gaussian Splatting SLAM~\cite{matsuki2024gaussian} proposed a novel Gaussian insertion and pruning strategy, enabling it to work not only with RGB-D but also with monocular cases. Additionally, PhotoSLAM~\cite{photo} leveraged explicit geometric features like ORB for localization and introduced a hyper primitives map for photometric feature mapping. However, most of the Gaussian Splatting SLAM approaches have focused on static scenes, overlooking dynamic environments. To address this issue, we propose Dynamic Gaussian Splatting SLAM (DGS-SLAM), which filters out dynamic elements across the entire Gaussian Splatting SLAM system.
\section{method}
\label{sec:method}

The overview of our DGS-SLAM is illustrated in Figure~\ref{fig:fig2}. Similar to existing Gaussian Splatting SLAM~\cite{matsuki2024gaussian,photo}, our system is composed of a frontend tracking and a backend mapping process after Gaussian map initialization. DGS-SLAM simultaneously optimizes the camera pose $\{\textbf{T}\}_{k=1}^N, \textbf{T}_k \in \mathbb{SE}(3)$ and reconstructs 3D Gaussian splatting with dynamic elements removed from a sequence of RGB-D frame inputs $\{I_k,D_k\}_{k=1}^N$.

\subsection{Gaussian Splatting}

Our dynamic SLAM system utilizes Gaussian splatting as the map representation. Each anisotropic Gaussian $\mathcal{G}^i$ is parameterized by its RGB color $c^i$, mean position $\mu^i \in \mathbb{R}^3$, covariance $\Sigma^i$, and opacity $o^i \in [0,1]$. We simplify each Gaussian to view-independent color, removing spherical harmonics (SHs). The Gaussian equation for a 3D point $x \in \mathbb{R}^3$ is as follows:
\begin{equation}
    g(x) = o\exp\left(-\frac{\lVert x-\mu  \rVert^2}{2r^2}\right).
\end{equation}

Following 3D Gaussian Splatting~\cite{3dgs}, each 3D Gaussian rasterizes into 2D splats, enabling gradient flow for scene reconstruction and pose estimation. We project $m$ Gaussians which are sorted in order of depth, and blend them into the color of pixel $p$:
\begin{equation}
    C(p) = \sum^{m}_{i=1}g_i(p)c_i\prod^{i-1}_{j=1}(1-g_i(p)), 
\label{eq:eq2}
\end{equation}

During rasterization, the mean position $\mu^{2D}$ and covariance $\Sigma^{2D}$ projected from 3D gaussian in pixel-space are as follows:
\begin{equation}
    \mu^{2D} = K\mathbf{T}\mu d^{-1},\, \Sigma^{2D} = J\mathbf{T}_{\text{rot}}\Sigma\mathbf{T}_{\text{rot}}^{T}J^T, 
\end{equation}
where $ d=(\mathbf{T}\mu)_z$ is the distance from the camera to the Gaussian and $K$ is the calibrated camera intrinsic parameter.

For the depth rendering of pixel $p$, we follow a similar to equation~\ref{eq:eq2}:
\begin{equation}
    D(p) = \sum^{m}_{i=1}g_i(p)d_i\prod^{i-1}_{j=1}(1-g_j(p)),\,
\end{equation}

\subsection{Pose Tracking}

During the tracking process in the frontend, the pose $T_k$ is optimized by minimizing the difference between each input frame $C_k,D_k$ and the rendered result $\hat{C}(T_k),\hat{D}(T_k)$ from the predicted pose $T_k$. However, it is necessary to filter out dynamic elements to accurately optimize pose estimation. For each input frame $k$, we utilize an off-the-shelf online instance video module~\cite{tracking} to obtain mask $M_{\text{seg}}^k$ of dynamic objects. Additionally, we generate opacity mask $M_{\text{opacity}}^k$ through opacity checks to filter out empty spaces before mapping. We combine the opacity mask and the segmentation mask to generate the overall tracking mask $M_{\text{tracking}}^k$ as:
\begin{equation}
    \widehat{M}_{\text{tracking}}^k = \widehat{M}_{\text{seg}}^k \otimes \left(\sum^{m}_{i=1}g_i(p)\prod^{i-1}_{j=1}(1-g_j(p)) \, > \tau_{\text{opacity}}\right).
 \end{equation}
 where $\tau_{\text{opacity}}$ is the threshold value to determine the unmapped region, which is set as 0.95.

We track the camera pose of frame $k$ by minimizing the following loss:
\begin{equation}
\begin{split}
    L_{\text{tracking}}^k &= \widehat{M}_{\text{tracking}}^k\,\left( \alpha L_{\text{color}}^k + (1-\alpha)L_{\text{depth}}^k\right),
\end{split}
\end{equation}
where $L_{\text{color}}^k$ is the photometric residual $\lVert \hat{C}_k(\mathcal{G},T_k) - C_k\rVert_1$ between rendered image and the input image, and $L_{\text{depth}}^k$ is the depth residual $\lVert \hat{D}_k(\mathcal{G},T_k) - D_k\rVert_1$. 

\subsection{ Loop-aware Keyframe Management}

\textbf{Keyframe selection}
After optimizing the camera pose $\mathbf{T}_k$ of the input frame $k$, we form a window $\mathcal{W}$ with keyframes to jointly optimize the Gaussians $\mathcal{G}$ and camera poses $\mathbf{T}_k,k\in\mathcal{W}$. The criteria for keyframe selection are Gaussian covisibility, relative pose $\mathbf{T}_{ij}$ from the last keyframe $j$, and the unique keyframe ID of the Gaussians. For Gaussian covisibility, we measure the intersection over union (IoU) of visible Gaussians between the current frame $i$ and the last keyframe $j$. A frame is registered if the IOU of covisibility $ \frac{|\mathcal{G}^v_i \cap \mathcal{G}^v_j|}{|\mathcal{G}^v_i \cup \mathcal{G}^v_j|}$ falls below a certain threshold $\sigma_{\text{IoU}}$. And for the relative pose $\mathbf{T}_{ij}$, when not only the transition but rotation difference of the relative pose exceeds a defined value, the current frame is the keyframe. Additionally, to maintain a consistent window length, keyframes are removed based on similar criteria.

\textbf{Loop-aware keyframe insertion} When the window $\mathcal{W}$ for joint optimization is managed solely based on covisibility and the relative pose with the last keyframe, removed keyframes can no longer affect the global map. This disrupts the consistency of the global Gaussian map. To address this, we introduce Loop-aware keyframe insertion, assigning the ID of the frame where it is generated to each Gaussian as $k \mapsto \text{ID}(\mathcal{G}_i)$. The unique keyframe ID of the visible Gaussian $\mathcal{G}^v_{k_c}$ in the current frame $k_c$ is key to identifying loops in past keyframes. The portion of these identified loop keyframes $k_l$ is re-included in the window and jointly optimized with the current keyframes as:
\begin{equation}
     \mathcal{W}=\mathcal{W}\cup\{k_l|k_l = \text{ID}(\mathcal{G}^v)\;\text{for}\; \mathcal{G}^v\;\text{in}\;\mathcal{G}^v_{k_c}\}.
\end{equation}

\subsection{Mapping process}
\textbf{Gaussian Insertion \& Pruning}

Before the camera tracking and mapping process begins, we insert and initialize the 3D Gaussians using the first frame input, $C_{k=0}, D_{k=0}$. To initialize the scene more accurately, we align the estimated depth $D_{\text{align}} = aD_{\text{est}}(C_{k=0}) + b$  with the ground truth depth $D_{t=0}$. Filling depth holes aid in completing the initial scene densely. Additionally, we define the position $\mu_{3D}$ of 3D Gaussian by unprojecting the pixels $p$ which are not on the invalid region of dynamic elements:
\begin{equation}
    \mu_{3D} =  \mathbf{T}^{-1}\cdot (K^{-1}p\cdot D(p)), \,\text{where}\; p \in M_{\text{seg}}. 
\label{eq:gs_mean}
\end{equation}

Initial Gaussians $\mathcal{G}_{init}$ are optimized by the corresponding loss $L_{\text{init}} = M_{\text{seg}, k=0}(\alpha L_{\text{color}} + (1-\alpha)L_{\text{depth})}$. Afterward, every time a new keyframe is added, 3D Gaussians are inserted according to Equation~\ref{eq:gs_mean}. For memory efficiency, Gaussians are generated more sparsely than initialization, and the scale 
 of each Gaussian is defined in proportion to the median depth, excluding invalid depths and dynamic regions. We primarily follow the Gaussian densification and pruning strategies as \cite{3dgs,matsuki2024gaussian}.

\textbf{Robust Mask Generation}

Online segmentation in real-world environments is not perfect. Therefore, to ensure photometric and geometric consistency, we additionally generate masks for outliers during the window optimization process. Similar to the approach proposed in \cite{refusion,robustnerf}, we sort the phometric residuals $ R_k = \lVert \hat{C}(\mathcal{G},T_k) - \Bar{C_k}\rVert_1$ between the rendered image and the input image of the keyframe $k$, and represent them as a histogram $\mathcal{H}(R_k)$. We assume that pixels with photometric residuals below a certain percentile $\tau_{\text{robust}}$ of the residual histogram are inliers. The residual threshold $\epsilon_{\text{robust}}$, which distinguishes inlier pixels from outlier pixels, is defined as follows:
\begin{equation}
    \epsilon_{\text{robust}} = \min \left\{ \epsilon \mid \sum_{i=0}^{\epsilon} \mathcal{H}_i \geq \tau_{\text{robust}} \cdot \sum_{i=0}^{\infty} \mathcal{H}_i \right\}
\end{equation}

To further avoid incorrect masking in high-frequency regions, we apply smoothing using a normalized kernel $\mathbf{W}$.
\begin{equation}
    M_{\text{robust}}^k = \mathds{1}\{ \left(\mathds{1}\{R_k > \epsilon_{\text{robust}}\} \circledast \textbf{W} \right)>0.5\}
\end{equation}

To gradually update the photometric residual histogram, The histogram used to determine outlier pixels is iteratively updated, as follows:
\begin{equation}
    \mathcal{H}^k_t = (1-\gamma) \cdot \mathcal{H}^k_{t-1} + \mathcal{H}(R_k)).
\end{equation}

\textbf{Window Optimization} In the backend mapping process, we optimize the 3D Gaussians using the keyframes selected through both window management and loop-aware selection. During the window optimization, we jointly optimize both the scene representation and the camera poses. Ultimately, we minimize the following loss,
\begin{equation}
    \min_{\substack{\mathbf{T}^k, \mathcal{G}, \\ \forall k \in \mathcal{W}}}
    \sum_{\forall k \in \mathcal{W}} M_{\text{window}}^k (\alpha L^{k}_{pho}+ (1-\alpha)L^{k}_{Depth}) + \lambda_{iso} L_{iso},
\end{equation}

where $M_{\text{window}}^k$ is the mask for window optimization obtained as $M_{\text{seg}}^k \otimes (1-M_{\text{robust}}^k)$, and $L_{iso}$ is an isotropic regularization term. This isotropic regularization aids in reducing artifacts in mapping 3D Gaussian Splatting.

\section{Experiments}

\textbf{Datasets}
We evaluated our approach on two prominent dynamic datasets: the TUM RGB-D dataset~\cite{tumbenchmark} and the Bonn RGB-D dataset~\cite{refusion}. Both datasets were captured in indoor environments using a handheld device, providing RGB images, depth maps, and ground truth trajectories. 

\textbf{Metrics}
To evaluate the quality of novel view synthesis for the map, we report the representative photometric metrics: Peak Signal-to-Noise Ratio (PSNR), Structural Similarity Index Measure (SSIM), and Learned Perceptual Image Patch Similarity (LPIPS). For assessing the rendering quality of dynamic scenes, we masked out the pixels corresponding to dynamic objects in black during evaluation. To measure camera tracking accuracy, we adopted the RMSE and standard deviation (STD) of the Absolute Trajectory Error (ATE)~\cite{tumbenchmark} of the keyframes.

\textbf{Implementation details}
Our DGS-SLAM experiments were conducted on a workspace equipped with a TITAN RTX and a 3.4GHz AMD Ryzen 7 3800XT. For the experiments, the loss parameters were set as follows: $\alpha=0.9, \delta_{iso} =0.1$, the robust threshold $\sigma_{robust}$ for robust mask estimation is 0.9, and the smoothing kernel size is 7. Additionally, the opacity threshold $\sigma_{opacity}$ for tracking is set to 0.95 and IoU threshold $\sigma_{\text{IoU}}$ to 0.9. We use the Adam optimizer~\cite{kingma2014adam} for camera optimization, with a learning rate of 0.003 for rotation and 0.001 for translation. The hyperparameters for the 3D Gaussians followed the same settings as~\cite{3dgs}. For the segmentation mask, we leverage Track Anything~\cite{tracking, sam,grounding}, an open-vocabulary video segmentation module that operates online. In particular, we implemented a lightweight tiny version to save computation time during experiments.
\setlength{\tabcolsep}{3.5pt}
\begin{table*}[!t]
\vspace{-2mm}
\caption{Novel View Synthesis Results as a mapping quality on several dynamic sequences in the \textit{TUM RGB-D} dataset.}
\vspace{-4mm}
\begin{center}
\footnotesize
\resizebox{1.0\linewidth}{!}{
\begin{tabular}{l|ccc|ccc|ccc|ccc|ccc}
\toprule

 &  \multicolumn{3}{c|}{\texttt{f3/wk\_xyz}} & \multicolumn{3}{c|}{\texttt{f3/wk\_hf}} & \multicolumn{3}{c|}{\texttt{f3/wk\_st}} & \multicolumn{3}{c|}{\texttt{f3/st\_hf}}  & \multicolumn{3}{c}{\texttt{Avg.}} \\

 & PSNR↑ & SSIM↑ & LPIPS↓  & PSNR↑ & SSIM↑ & LPIPS↓ & PSNR↑ & SSIM↑ & LPIPS↓ & PSNR↑ & SSIM↑ & LPIPS↓ & PSNR↑ & SSIM↑ & LPIPS↓\\
\midrule
SplaTAM \cite{keetha2024splatam} & 14.54 & 0.539 & 0.480 & 13.52 & 0.507 & 0.517 & 17.90 & 0.802 & 0.270 & 16.25 & 0.680 & 0.384 & 15.55 & 0.633 & 0.413 \\
MonoGS \cite{matsuki2024gaussian}  & 14.41 & 0.535 & 0.391 & 14.23 & 0.542 & 0.457 & 16.81 & 0.714 & 0.252 & 18.11 & 0.692  & 0.333  & 15.89  & 0.621  & 0.358 \\
GS-ICP \cite{ha2024rgbd}& 16.92 & 0.694 & 0.356 & 16.36 & 0.671 & 0.392 & 18.30 & 0.727 & 0.311 & 18.12 & 0.728 & 0.293 & 17.42 & 0.705 & 0.338 \\
DGS-SLAM \textbf{(Ours)} & \textbf{20.48} & \textbf{0.798} & \textbf{0.173} & \textbf{20.00} & \textbf{0.774} & \textbf{0.229} & \textbf{22.89} & \textbf{0.897} & \textbf{0.089} & \textbf{19.16} & \textbf{0.763} & \textbf{0.251} & \textbf{20.63} & \textbf{0.807} & \textbf{0.186}\\
\bottomrule 
\end{tabular}
}
\label{table:rendering}
\vspace{-2mm}
\end{center}
\end{table*}

\setlength{\tabcolsep}{5pt}
\begin{table*}[!t]
\vspace{-2mm}
\caption{Camera tracking results on dynamic and static scenes in the \textit{TUM RGB-D} dataset. The units for ATE and S.D are in cm.}
\vspace{-4mm}
\begin{center}
\footnotesize
\begin{tabular}{l|c|cc|cc|cc|cc|cc|cc|cc}
\toprule
\multirow{2}{*}{Methods} &\multirow{2}{*}{Dense} &\multicolumn{8}{c|}{Dynamic} &\multicolumn{4}{c|}{Static} &\multicolumn{2}{c}{\multirow{2}{*}{Avg.}}\\
&  & \multicolumn{2}{c}{\texttt{f3/wk\_xyz}} & \multicolumn{2}{c}{\texttt{f3/wk\_hf}} & \multicolumn{2}{c}{\texttt{f3/wk\_st}} & \multicolumn{2}{c|}{\texttt{f3/st\_hf}}  & \multicolumn{2}{c}{\texttt{f1/xyz}} & \multicolumn{2}{c|}{\texttt{f1/rpy}} \\
\midrule
\textit{Traditional SLAM methods}  &  & \textit{ATE} &\textit{S.D.}  & \textit{ATE} &\textit{S.D.}   & \textit{ATE} &\textit{S.D.}  &  \textit{ATE} &\textit{S.D.}   & \textit{ATE} &\textit{S.D.}  & \textit{ATE} &\textit{S.D.} & \textit{ATE} &\textit{S.D.}\\
ORB-SLAM3 \cite{orbslam3} &  & 28.1 & 12.2 & 30.5 & 9.0 &2.0 & 1.1 & \textbf{2.6}  & 1.6 &\textbf{1.1}  &0.6 &2.2  &1.3 &11.1  &4.3\\
DVO-SLAM \cite{dvoslam} & \cmark & 59.7 & - & 52.9 & - &21.2 & - & 6.2  & - &\textbf{1.1}  &- &\textbf{2.0}  &- &22.9  &-\\
DynaSLAM \cite{dynaslam} &  & \textbf{1.7} & -  & \textbf{2.6} & - & \textbf{0.7} & -  & 2.8 & - & -  &- &-  &- &\textbf{2.0}  &-\\
ReFusion \cite{refusion} & \cmark & 9.9 & -  & 10.4 & -  &1.7 & -  & 11.0 & - &- &- &-  &- &8.3  &-\\
\midrule
\textit{Radiance-Field SLAM methods} &  &\textit{ATE} &\textit{S.D.}  & \textit{ATE} &\textit{S.D.}   & \textit{ATE} &\textit{S.D.}  &  \textit{ATE} &\textit{S.D.}   & \textit{ATE} &\textit{S.D.}  & \textit{ATE} &\textit{S.D.} & \textit{ATE} &\textit{S.D.}\\
NICE-SLAM \cite{nice_slam} & \cmark &113.8  & 42.9 & X & X  &88.2  & 27.8 & 45.0 &14.4 & 4.6 & 3.8 &3.4  & 2.5 & 51 & 18.3\\
Vox-Fusion \cite{voxfusion} & \cmark &146.6   & 32.1  & X & X &109.9  & 25.5 & 89.1 & 28.5 &1.8  & 0.9 &4.3   & 3.0 &70.4 &18\\
Co-SLAM \cite{co-slam} & \cmark &51.8 & 25.3  & 105.1 & 42.0 &49.5 & 10.8  & 4.7 & 2.2 & 2.3 & 1.2 &3.9 & 2.8 &36.3 &14.1\\
ESLAM \cite{eslam} & \cmark &45.7 & 28.5 & 60.8 & 27.9 &93.6 & 20.7  & 3.6 & 1.6 & 1.1 & 0.6 &\textbf{2.2} & \textbf{1.2} &34.5 &13.5\\
SplaTAM \cite{keetha2024splatam} & \cmark & 134.4 & 32.1 & 746.1 & 250.5 & 97.8 & 26.9 & 14.1 & 6.8 & \textbf{1.0} & 0.5 & 2.6 & 1.3 & 166.0 & 52.9 \\
MonoGS \cite{matsuki2024gaussian} & \cmark & 73.4 & 20.1 & 65.6 & 24.8 & 5.5 & 3.0 & \textbf{2.7} & \textbf{1.5}  & \textbf{1.0} & \textbf{0.4} & 2.5 & 1.3 & 37.7 & 25.1 \\
GS-ICP SLAM \cite{ha2024rgbd} & \cmark & 70.5 & 45.1 & 73.9 & 34.1 & 98.2 & 24.1 & 9.9 & 3.7 & 1.4 & 0.7 & 3.2 & 2.8 & 42.9 & 18.4 \\
RoDyn-SLAM \cite{jiang2024rodyn} & \cmark & 8.3 & 5.5 & 5.6 & \textbf{2.8} & 1.7  & 0.9 & 4.4 & 2.2 & 1.5 & 0.8 & 2.8 & 1.5 & 4.1 &2.3\\
DGS-SLAM \textbf{(ours)} & \cmark & \textbf{4.1} & \textbf{2.2} & \textbf{5.5} & \textbf{2.8} & \textbf{0.6}  & \textbf{0.2} & 4.1 & 1.6  & 1.2 & 0.6 & 2.4 & 1.3 & \textbf{3.0} & \textbf{1.5} \\
\bottomrule 
\end{tabular}
\label{table:tum_tracking}
\vspace{-6mm}
\end{center}
\end{table*}
\begin{figure}[htbp!]
\centering
\includegraphics[width=\columnwidth]{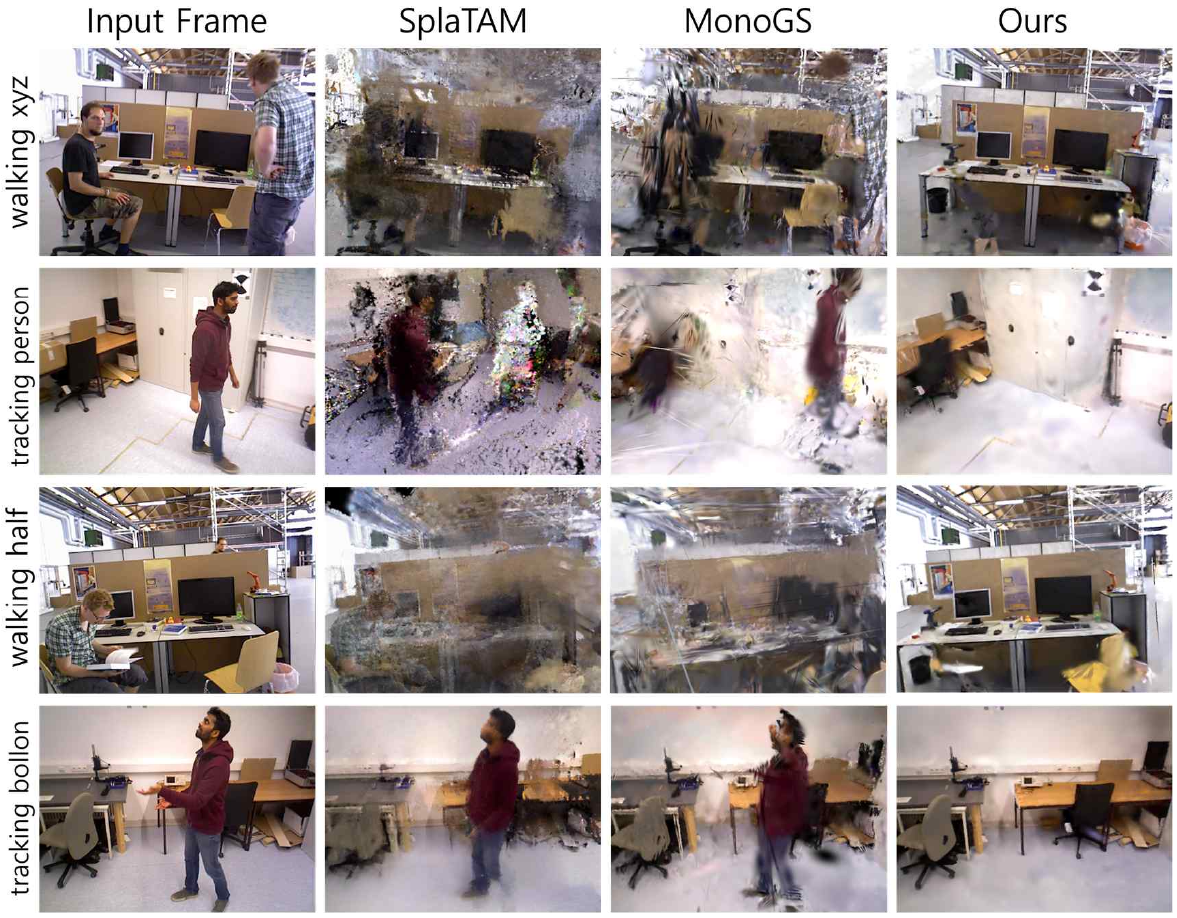}
\caption{Comparison of rendered results from state-of-the-art Gaussian Splatting SLAM approaches based on the estimated input frame poses.}
\vspace{-15pt}
\label{fig:rendering_compare} 
\end{figure}

\subsection{Evaluation of Mapping} 
To demonstrate the mapping performance of our DGS-SLAM in dynamic environments, we evaluated its novel view synthesis performance as a qualitative result. We evaluated rendering quality by averaging the differences between the rendered images and the ground truth images across all frames. We show a comparison of rendering quality with Gaussian Splatting-based SLAM methods, SplaTAM~\cite{keetha2024splatam} and MonoGS~\cite{matsuki2024gaussian}. As shown in Table~\ref{table:rendering}, our proposed DGS-SLAM achieves the best performance among Gaussian Splatting SLAMs on the dynamic scenes of the TUM dataset. In Figure~\ref{fig:rendering_compare} , we qualitatively compare the rendered image from reconstructed Gaussian maps, showing how our method is robust in a dynamic environment compared to other GS-SLAM approaches. In dynamic scenes, existing GS-based SLAM systems generate Gaussians for moving elements, leading to photometric inconsistencies and failing to accurately reconstruct the scene.

\subsection{Evaluation of Camera Tracking}

To evaluate the camera tracking performance in dynamic environments, we compared our method with radiance field based SLAM methods~\cite{nice_slam,co-slam,keetha2024splatam,matsuki2024gaussian,ha2024rgbd} and traditional SLAM methods~\cite{orbslam3,dvoslam,refusion}, including SLAM methods~\cite{dynaslam,teed2021droid,jiang2024rodyn} specifically designed for dynamic environments. In Table~\ref{table:tum_tracking}, we present the results for four dynamic scenes from the TUM dataset. We achieved superior results through the advantages of using Gaussian splatting as a map representation, along with a novel robust mask and a novel keyframe management strategy. While existing Gaussian Splatting-based SLAMs often fail in dynamic scenes, our method demonstrates accurate camera tracking. Furthermore, compared to RoDyn-SLAM~\cite{jiang2024rodyn}, a NeRF-based dynamic SLAM, our approach achieves more accurate results in most of the scenes. Our approach even outperforms traditional dynamic SLAMs in $\textit{walking\_xyz}$ scene. Table~\ref{table:bonn_tracking} presents the camera tracking results on the Bonn RGB-D dataset~\cite{refusion}. The Bonn dataset is more complex and captured in larger scenes with various dynamic movements. In Bonn dataset, our DGS-SLAM demonstrates superior performance compared to other radiance-field based SLAM methods, outperforming some traditional sparse SLAM approaches.

\begin{figure*}[htbp!]
\centering
\includegraphics[width=0.9\textwidth]{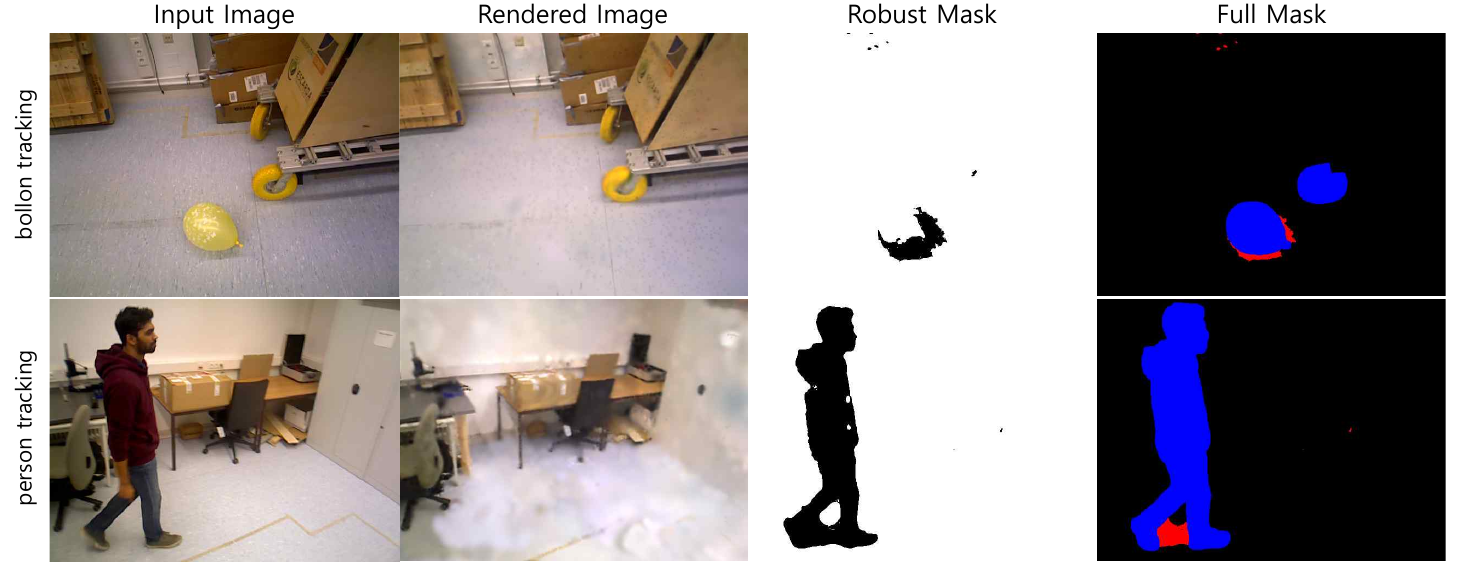}
\vspace{-5pt}
\caption{Visualization of robust mask generation. From right to left: the input image, rendered image, robust mask, and full mask. In the full mask, blue represents the semantic segmentation mask, and red indicates the robust mask.}
\label{fig:robust} 
\end{figure*}
\setlength{\tabcolsep}{3.5pt}
\begin{table*}[htbp]
\vspace{-2mm}
\caption{Camera tracking results on dynamic scenes in the \textit{BONN RGB-D} dataset. The units for ATE and S.D are in centimeters (cm).}
\vspace{-4mm}
\begin{center}
\begin{tabular}{l|c|cc|cc|cc|cc|cc|cc|cc}
\toprule
\multicolumn{1}{l}{Methods} & \multicolumn{1}{c}{Dense} & \multicolumn{2}{c}{\texttt{balloon}} & \multicolumn{2}{c}{\texttt{balloon2}} & \multicolumn{2}{c}{\texttt{ps\_track}} & \multicolumn{2}{c}{\texttt{ps\_track2}} & \multicolumn{2}{c}{\texttt{ball\_track}} & \multicolumn{2}{c}{\texttt{mv\_box2}}  & \multicolumn{2}{c}{Avg.} \\
\midrule
\textit{Traditional SLAM methods} &  & \textit{ATE} &\textit{S.D.}  & \textit{ATE} &\textit{S.D.}   & \textit{ATE} &\textit{S.D.}  &  \textit{ATE} &\textit{S.D.}   & \textit{ATE} &\textit{S.D.}  & \textit{ATE} &\textit{S.D.} & \textit{ATE} &\textit{S.D.}\\
ORB-SLAM3 \cite{orbslam3} & & 5.8 & 2.8  & 17.7 & 8.6 & 70.7 & 32.6  & 77.9 & 43.8 & \textbf{3.1} &1.6 &\textbf{3.5} &1.5 & 29.8 &15.2\\
Droid-VO \cite{teed2021droid} &\cmark & 5.4 & -  & 4.6 & -  &21.34 & -  & 46.0 & - &8.9 &- &5.9 &- &15.4 &-\\
DynaSLAM \cite{dynaslam} &  & \textbf{3.0} & -  & \textbf{2.9} & - & \textbf{6.1} & -  &\textbf{7.8} & - & 4.9 &-& 3.9 &- & \textbf{4.8} &-\\
ReFusion \cite{refusion} &\cmark & 17.5 & -  & 25.4 & -  &28.9 & -  & 46.3 & - &30.2 & - & 17.9 &- & 27.7 &-\\
\midrule
\textit{Radiance-Field SLAM methods} &  & \textit{ATE} &\textit{S.D.}  & \textit{ATE} &\textit{S.D.}   & \textit{ATE} &\textit{S.D.}  &  \textit{ATE} &\textit{S.D.}   & \textit{ATE} &\textit{S.D.}  & \textit{ATE} &\textit{S.D.} & \textit{ATE} &\textit{S.D.}\\
NICE-SLAM \cite{nice_slam} &\cmark &X  & X &66.8 & 20.0  &54.9  & 27.5 & 45.3 &17.5 & 21.2  & 13.1 &31.9  &13.6 &44.1 &18.4\\
Vox-Fusion \cite{voxfusion} &\cmark &65.7  & 30.9  & 82.1 & 52.0 &128.6  & 52.5 & 162.2 & 46.2  &43.9  & 16.5 & 47.5  & 19.5 &88.4 &36.3\\
Co-SLAM \cite{co-slam} &\cmark &28.8 & 9.6  & 20.6 & 8.1 &61.0 & 22.2  & 59.1 & 24.0 & 38.3 & 17.4 & 70.0 & 25.5 &46.3 & 17.8\\
ESLAM \cite{eslam} &\cmark &22.6 & 12.2  & 36.2 & 19.9 &48.0 & 18.7  & 51.4 & 23.2  & 12.4 & 6.6 & 17.7 & 7.5 & 31.4 & 14.7\\
SplaTAM \cite{keetha2024splatam} & \cmark & 35.7 & 14.1 & 36.4 & 17.4 & 124.8 & 36.5 & 163.0 & 51.3 & 12.8 & 16.8 & 17.9 & 9.3 & 65.1 & 24.2 \\
MonoGS \cite{matsuki2024gaussian} & \cmark & 33.2 & 16.4  & 26.5 & 14.2 & 63.2  & 29.0 & 47.2 & 15.4 & 4.3 & 2.2 & 22.9 & 12.4 & 32.9 & 14.2 \\
GS-ICP SLAM \cite{ha2024rgbd} & \cmark &  43.8 & 16.0 & 42.1 & 19.1 & 92.8 & 42.3 & 44.7 & 20.3 & 27.9 & 17.4 & 24.8 & 11.5 & 31.3 & 14.2 \\
RoDyn-SLAM \cite{jiang2024rodyn} &\cmark & 7.9 & 2.7 & 11.5 & 6.1 & 14.5 & 4.6 & 13.8 & \textbf{3.5} & 13.3 & 4.7 & 12.6  & 4.7 &12.3 & 4.4\\
DGS-SLAM \textbf{(Ours)} &\cmark & \textbf{2.9} & \textbf{0.8} &\textbf{6.0} &\textbf{2.8}  & \textbf{9.8}  & \textbf{4.1} & \textbf{11.1} & 3.9 & \textbf{5.6} &  \textbf{2.8}  & \textbf{8.8} &  \textbf{3.8} & \textbf{7.3} & \textbf{3.0} \\
\bottomrule 
\end{tabular}
\label{table:bonn_tracking}
\vspace{-5mm}
\end{center}
\end{table*}

\subsection{Ablation study}
\textbf{Ablative Analysis}
By evaluating our framework by removing each component individually, we demonstrate the contribution of each proposed method. Table~\ref{tab:ablation_study} presents the RMSE ATE and the mean STD of camera tracking results across six scenes from the Bonn dataset. When loop-aware keyframe selection was removed, the performance in large-scale environments declined, and the robust mask contributed to stability in pose tracking.

\setlength{\tabcolsep}{5pt}
\begin{table}[!t]

\caption{Ablation study of our proposed methods. }
\vspace{-4mm}
    \centering
    \footnotesize
\centering
\begin{center}
        \begin{tabular}{c|c|c|c|c}
            \toprule
            & w/o Both & w/o Robust & w/o Loop  & DGS-SLAM \\ \midrule
            ATE (cm) $\downarrow$& 9.036 & 8.343 & 8.008 & \textbf{7.322}  \\
            STD (cm) $\downarrow$& 3.557 & 3.802 & 3.164 & \textbf{3.015} \\
            \bottomrule
        \end{tabular}
\label{tab:ablation_study}
 \end{center}
 \vspace{-8mm}
 \end{table}
\textbf{Robust Mask Visualization} We visualize the robust mask generated during the window optimization in Figure~\ref{fig:robust}. The results of the generated mask demonstrate that our method not only extracts masks corresponding to semantic information but also artifacts like shadows created by dynamic objects. Our robust mask generation is achieved through iteratively updating photometric residuals, leveraging the high-level rendering capabilities of Gaussian splatting.

\setlength{\tabcolsep}{8pt}
\begin{table}[!h]
\vspace{-3mm}
\caption{Time analysis}
\vspace{-4mm}
    \centering
    \footnotesize
\centering
\begin{center}
    \setlength{\tabcolsep}{0.5cm}
        \begin{tabular}{c|c|c}
            \toprule
             & Total Time (Sec) & FPS (Hz)  \\ \midrule
            MonoGS~\cite{matsuki2024gaussian} &   776.4  &  1.07   \\
            Ours   &  519.0 &  1.60 \\
            \bottomrule
        \end{tabular}
\label{tab:time_analysis}
 \end{center}
 \vspace{-5mm}
 \end{table}

\textbf{Time Analysis}
We measured the total SLAM process time for the $\textit{fr3/wk\_xyz}$ sequence from the TUM RGB-D dataset in our workspace, excluding the time spent on semantic segmentation. Our framework took 519 seconds for the total processing time, and when divided by the total number of frames, it achieved a performance of 1.60 FPS as shown in Table~\ref{tab:time_analysis}. In comparison, the baseline Gaussian Splatting SLAM~\cite{matsuki2024gaussian} showed only 1.07 FPS due to the presence of dynamic elements. Note that it reaches 1.60 FPS in static conditions, similar to our framework. This demonstrates that our framework effectively processes the dynamic SLAM pipeline with minimal additional computation.

\section{Conclusion}

We present DGS-SLAM, the first dynamic Gaussian Splatting SLAM. Our approach handles dynamic elements not only during the tracking and mapping stages but also across the entire Gaussian Splatting SLAM system, enabling robust pose tracking and reconstruction of the Gaussian splatting. Additionally, to achieve precise localization and mapping performance, we introduce a robust mask generation method that leverages photometric residuals. Furthermore, our loop-aware keyframe management accounts for loops with past frames, ensuring consistency to the Gaussian map. Our method demonstrates state-of-the-art performance among radiance-field-based SLAM approaches on two representative dynamic datasets. 

\newpage
\addtolength{\textheight}{-12cm}   
\bibliographystyle{IEEEtran}
\bibliography{ref}

\end{document}